\pdfoutput=1

\documentclass[11pt]{article}
\usepackage{emnlp2021}
\usepackage{times}
\usepackage{latexsym}
\usepackage{graphicx}
\usepackage{multirow}
\usepackage{booktabs}
\usepackage{subfig}
\usepackage{multicol}

\usepackage[T1]{fontenc}
\usepackage[utf8]{inputenc}
\usepackage{array}

\usepackage{microtype}

\newcommand\sys{\emph{Tribrid}}
\newcommand\syspos{\emph{Tribrid$_{pos}$}}
\newcommand\syslog{\emph{Tribrid$_{l}$}}
\newcommand\sysdist{\emph{Tribrid$_{d}$}}
\newcommand\sysweak{\emph{Tribrid$_{w}$}}
\newcommand\bertbase{BERT$_{base}$}
\newcommand\bertbaseneg{BERT$_{ba}$/N}
\newcommand\stancy{STANCY}
\newcommand\stancyneg{STANC/N}
\newcommand\claim{\ensuremath{C}}
\newcommand\pers{\ensuremath{P}}
\newcommand\negpers{\ensuremath{NP}}
\newcommand\claimpersrep{\ensuremath{\Sigma_{cp}}}
\newcommand\claimrep{\ensuremath{\Sigma_c}}
\newcommand\persrep{\ensuremath{\Sigma_p}}
\newcommand\negpersrep{\ensuremath{\Sigma_{\neg p}}}
\newcommand\lpos{\ensuremath{\lambda_s}}
\newcommand\lneg{\ensuremath{\lambda_o}}
\newcommand\loss{\ensuremath{\mathcal{L}}}
\newcommand\losscos{\ensuremath{\mathcal{L}_c}}
\newcommand\lossent{\ensuremath{\mathcal{L}_e}}
\newcommand\lossdistance{\ensuremath{\mathcal{L}_d}}
\newcommand\distsupport{\ensuremath{\delta^{+}}}
\newcommand\distoppose{\ensuremath{\delta^{-}}}
\newcommand\distpers{\ensuremath{\delta_{p}}}
\newcommand\distnegpers{\ensuremath{\delta_{\neg p}}}
\newcommand\support{\ensuremath{\mathsf{S}}}
\newcommand\oppose{\ensuremath{\mathsf{O}}}
\newcommand\abstain{\ensuremath{\mathsf{A}}}
\newcommand\classlog{\ensuremath{K^\tau}}
\newcommand\classdist{\ensuremath{\Lambda^\tau}}
\newcommand\classlogvar[1]{\ensuremath{K^{#1}}}
\newcommand\classdistvar[1]{\ensuremath{\Lambda^{#1}}}

\title{Tribrid: Stance Classification with Neural Inconsistency Detection}

\author{Song Yang\\
    Department of Computer Science\\
    Vrije Universiteit Amsterdam\\
    The Netherlands\\
  \texttt{s.yang@student.vu.nl} \\\And
  Jacopo Urbani\\
    Department of Computer Science\\
    Vrije Universiteit Amsterdam\\
    The Netherlands\\
  \texttt{jacopo@cs.vu.nl} \\}

\date{}

\begin{document}
\maketitle

\begin{abstract}

We study the problem of performing automatic stance classification on social
media with neural architectures such as BERT. Although these architectures
deliver impressive results, their level is not yet comparable to the one of
humans and they might produce errors that have a significant impact on the
downstream task (e.g., fact-checking). To improve the performance, we present a
new neural architecture where the input also includes automatically generated
negated perspectives over a given claim. The model is jointly learned to make
simultaneously multiple predictions, which can be used either to improve the
classification of the original perspective or to filter out doubtful
predictions. In the first case, we propose a weakly supervised method for
combining the predictions into a final one. In the second case, we show that
using the confidence scores to remove doubtful predictions allows our method to
achieve human-like performance over the retained information, which is still a
sizable part of the original input.

\end{abstract}

\section{Introduction}

The spreading of unverified claims on social media is an important problem that
affects our society at multiple
levels~\cite{vlachos_fact_2014,ciampaglia_computational_2015,hassan_claimbuster_2017}.
A valuable asset that we can exploit to fight this problem is the set of
perspectives that people publish about such claims. These perspectives reveal
the users' stance and this information can be used to help an automated
framework to determine more accurately the veracity of
rumors~\cite{10.1145/1963405.1963500,bourgonje-etal-2017-clickbait}.

The stance can be generally categorized either as supportive or opposing. For
instance, consider the claim ``\emph{The elections workers in Wisconsin
illegally altered absentee ballot envelopes}''. A tweet with a supportive
stance is ``\emph{The number of people who took part in the election in
Wisconsin exceeded the total number of registered voters}'' while one
with an opposed stance is ``\emph{An extra zero was added as votes
accidentally but it was quickly fixed after state officials noticed it}''.

Being able to automatically classify the stance is necessary for dealing with
the large volume of data that flows through social networks. To this end,
earlier solutions relied on linguistic features, such as n-grams, opinion
lexicons, and
sentiment~\cite{somasundaran2009recognizing,anand2011cats,hasan2013stance,sridhar_joint_2015}
while more recent methods additionally include features that we can extract from
networks like
Twitter~\cite{chen2016utcnn,lukasik-etal-2016-hawkes,sobhani_dataset_2017,kochkina2017turing}.
The current state-of-the-art relies on advanced neural architectures such as
BERT~\cite{devlin2018bert} and returns remarkable performance. For instance,
STANCY~\cite{popat2019stancy} can achieve an $F_1$ of 77.76 with the PERSPECTRUM
(PER) dataset against the $F_1$ of 90.90 achieved by
humans~\cite{chen2019seeing}.

Although these results are encouraging, the performance has not yet reached a
level that it can be safely applied in contexts where errors must be avoided at
all costs. Consider, for instance, the cases when errors lead to a
misclassification of news about a catastrophic event, or when they trigger wrong
financial operations. In such contexts, we argue that it is better that the AI
abstains from returning a prediction unless it is very confident about it. This
requirement clashes with the design of current solutions, which are meant to
``blindly'' make a prediction for any input. Therefore, we see a gap between the
capabilities of the state-of-the-art and the needs of some realistic use cases.

In this paper, we address this problem with a new BERT-based neural network,
which we call \sys{} (TRIplet Bert-based Inconsistency Detection), that is
designed not only to produce a reliable and accurate classification, but also to
test its confidence. This test is implemented by including a ``negated'' version
of the original perspective as part of the input, following the intuition that a
prediction is more trustworthy if the model produces the opposite outcome with
the negated perspective. If that is not the case, then the model is inconsistent
and the prediction should be discarded.

Testing the consistency of the model with negated perspectives is a task that
can be done simply by computing two independent predictions, one with the
original perspective and one with the negated one. However, this is suboptimal
because existing state-of-the-art methods are trained only with the principle
that supportive perspectives should be similar to their respecting claims in the
latent space~\cite{popat2019stancy} and the similarity might be sufficiently
high even if there are keywords (e.g., ``not'') that negate the stance. To
overcome this problem, we propose a new neural architecture that processes
\emph{simultaneously} both original and negated perspectives, using a siamese
BERT model and a loss function that maximises the distance between the two
perspectives. In this way, the model learns to distinguish more clearly
supportive and opposite perspectives.

To cope with the large volume of information that flows through social networks,
it is important that the negated perspectives are generated automatically or at
least with a minimal human intervention. To this end, two types of techniques
have been presented in the literature. One consists of attaching a fixed phrase
which negates the meaning~\cite{bilu2015automatic} while the other adds or
removes the first occurrences of tokens like ``not''
\cite{niu2018adversarial,camburu2020make}.  \sys{} implements this task in a
different way, namely using simple templates that negate both with keywords
(e.g., ``not'') and antonyms.

The prediction scores obtained with \sys{} can be used either to determine more
accurately the stance of the original perspective or to discard low-quality
predictions. We consider both cases: In the first one, we propose an approach
where multiple classifiers are constructed from the scores and a final weakly
supervision model combines them. In the second one, we propose several
approaches to establish the confidence and describe how to use them to discard
low-quality predictions. Our experiments show that our method is competitive in
both cases. For instance, in the second case, our approach was able to achieve a
$F_1$ of 87.43 on PER by excluding only 39.34\% of the perspectives. Moreover,
the score increased to 91.26 when 30\% more is excluded. Such performance is
very close to the one of humans and this opens the door to an application in
contexts where errors are very costly.

The source code and other experimental data can be found at
\url{https://github.com/karmaresearch/tribrid}.

\begin{figure*}[t]
    \centering
    \subfloat[\bertbase{}]{\label{fig:bert}\includegraphics[width=0.4\textwidth]{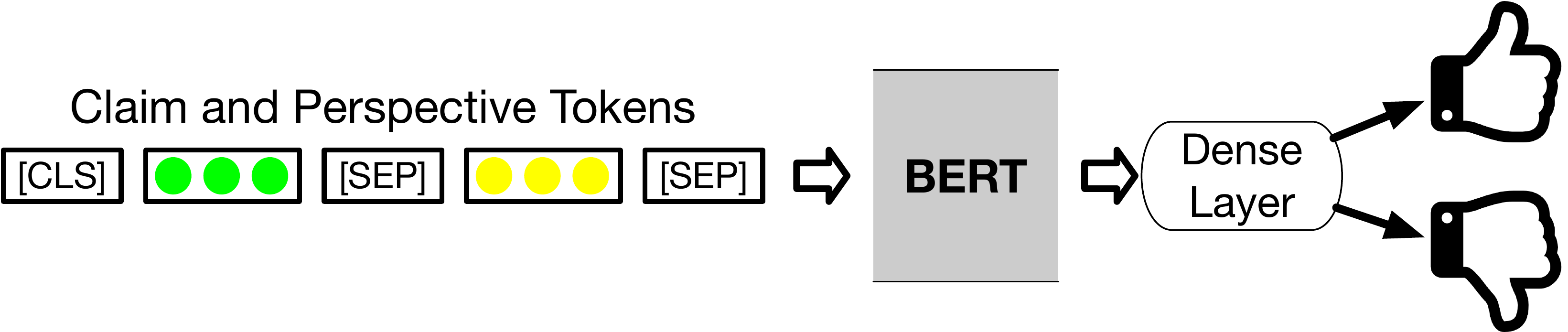}}
    \hfill
    \subfloat[STANCY]{\label{fig:stancy}\includegraphics[width=0.55\textwidth]{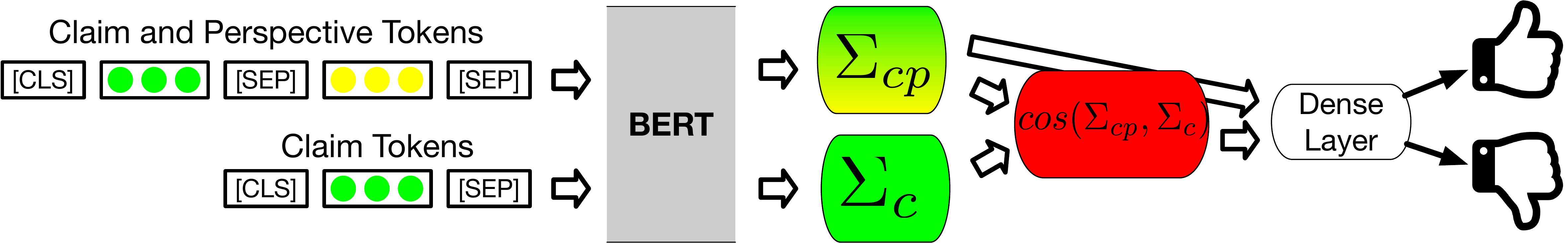}}
    \caption{BERT used for stance classification
        by~\citeauthor{chen2019seeing}~\shortcite{chen2019seeing}
    and~\citeauthor{popat2019stancy}~\shortcite{popat2019stancy}}
    \label{fig:bert_architectures}
\end{figure*}

\section{Background and Related Work}

\textbf{Stance Classification} aims to determine the stance of a input
perspective that supports or opposes another given claim. In earlier studies,
the research mainly focused on online debate posts using traditional
classification approaches \cite{thomas2006get, murakami2010support,
walker2012stance}. Afterwards, other approaches focused on spontaneous speech
\cite{levow2014recognition}, and on student essays \cite{faulkner2014automated}.
Thanks to the rapid development of social media, the number of studies on tweets
has increased substantially
\cite{rajadesingan2014identifying,chen2016utcnn,lukasik-etal-2016-hawkes,sobhani_dataset_2017,kochkina2017turing},
especially boosted by dedicated SemEval challenges~\cite{mohammad2016semeval,
kochkina2017turing} and benchmarks~\cite{bar-haim-etal-2017-stance,
chen2019seeing}.

Ealier methods for stance classification employed various traditional
classifiers, which include rule-based algorithms~\cite{anand2011cats};
supervised classifiers like SVM \cite{hasan2013stance}, naïve Bayes
\cite{rajadesingan2014identifying}, boosting \cite{levow2014recognition},
decision tree and random forest \cite{misra2013topic}, Hidden Markov Models
(HMM) and Conditional Random Fields \cite{hasan2013stance}; graph algorithms
such as MaxCut \cite{murakami2010support}, and other approaches such as Integer
Linear Programming \cite{somasundaran2009recognizing} and Probabilistic Soft
Logic \cite{sridhar2014collective}. Popular features include cue/topic words,
argument-related and sentiment/subjectivity features, and frame-semantic
features. Other features like tweets reply, rebuttal information, and retweets
are known to improve the performance \cite{sobhani2019exploring}.

In more recent work, the NLP community investigated on how to use deep neural
network to improve the performance. Some representatives are LSTM-based
approaches \cite{du2017stance, sun2018stance, wei2018multi}, RNN-based
approaches \cite{sobhani2019exploring, borges2019combining}, CNN approaches
\cite{wei2016pkudblab, zhang2017we} and more recently BERT-based approaches
\cite{chen2019seeing, popat2019stancy, schiller2020stance}.  Techniques like
attention mechanisms \cite{du2017stance}, memory networks
\cite{mohtarami2018automatic}, lexical features \cite{riedel2017simple,
hanselowski2018retrospective}, transfer learning and multi-task learning
\cite{schiller2020stance} can also improve the performance.

All these approaches focus on identifying the most effective method to achieve
the highest possible performance using syntactic and semantic features from the
input. In contrast, our goal is to improve the performance by injecting
background knowledge in the form of negated text, encouraging the model to
produce more consistent predictions. As far as we know, we are the first that
study this form of optimization to improve the performance of stance
classification.

\textbf{The generation of negated perspectives} can be viewed as an instance of
the broader problem of constructing adversarial examples, which is drawing more
attention in NLP research in recent years~\cite{zhang2020adversarial,
wang2019survey}. The main focus of adversarial generation is to change the text
(with character/words changes or removals) to train more robust models. Most of
the works focus on changes that do not alter the semantics of the original
input~\cite{belinkov2017synthetic,
xiao2017deceiving,iyyer2018adversarial,cheng2020seq2sick} while works that
negate the semantics are many fewer. In this context, some initial works have
manually constructed some small-scale tests \cite{isabelle2017challenge,
mahler2017breaking}. More recently,
\citeauthor{gardner2020evaluating}~\shortcite{gardner2020evaluating} suggested
that datasets should be perturbed by experts with small changes to the test
instances. In this way, empirical evaluations can test more accurately the true
linguistic capabilities of the models. In their evaluation, they selected PER,
one of the datasets that we also consider, and showed that models such as BERT
perform significantly worse on the perturbed dataset. Also
\citeauthor{ribeiro2020beyond}~\shortcite{ribeiro2020beyond} consider the
problem that accuracy on a held-out dataset may overestimate the performance on
a real scenario. To counter this, they propose a new methodology that involves
tests with certain perturbations (like negation), but they do not consider
stance classification. These works further motivate our effort to develop models
that are more consistent when presented with negated inputs.  Moreover, another
goal of our work is to use consistency as a proxy to measure uncertainty and to
discard low-quality predictions. A similar objective was pursued by
\citeauthor{kochkina2020estimating}~\shortcite{kochkina2020estimating} for the
problem of rumor verification.

Since manually creating additional test instances is time consuming, some works
propose automatic procedures to generate them.
\citeauthor{bilu2015automatic}~\shortcite{bilu2015automatic} proposes to add a
fixed phrase at the end (``but this is not true'') while
\citeauthor{niu2018adversarial}~\shortcite{niu2018adversarial} adds the token
``not'' before the first verb in the sentence or replaces it with its antonym.
Finally, \citeauthor{camburu2020make}~\shortcite{camburu2020make} suggests a
simpler alternative to remove the first occurrence of ``not''.  In contrast to
them, we use templates in the form of \emph{if-then} rules.

\section{Our Approach}

First, we provide a short description on how BERT has been used to achieve the
state-of-the-art for this problem (Section~\ref{sec:bert}). Then, we describe
our proposed neural network (Section~\ref{sec:nn}) and how we can interpret its
output to classify the stance (Section~\ref{sec:stance}). Finally, we discuss
the generation of negated perspectives using templates
(Section~\ref{sec:templates}).

\subsection{BERT Base and STANCY}
\label{sec:bert}

Our input is a sentence pair $\langle \claim, \pers \rangle$ where \claim{} is
the input claim and \pers{} is the perspective. In
2019,~\citeauthor{chen2019seeing} proposed to concatenate \claim{} and \pers{}
using two tokens \texttt{[CLS]} and \texttt{[SEP]} to delimit the claim and the
perspective, respectively, and to feed the resulting string to BERT (see
Figure~\ref{fig:bert}). We call this approach \bertbase{}. A few months
later,~\citeauthor{popat2019stancy}~\shortcite{popat2019stancy} proposed an
improvement based on the assumption that the latent representation of a
perspective should be similar if the perspective support the claim and vice
versa. The resulting network is called STANCY (see Figure~\ref{fig:stancy}). The
main idea is to compute a latent representation of the claim and perspective
(\claimpersrep{}) and one for the claim alone (\claimrep{}). The two are
compared with cosine similarity, denoted with $cos(\cdot)$, and passed to a
final dense layer that performs the classification.

\subsection{Tribrid: Neural Architecture}
\label{sec:nn}

\begin{figure*}
    \centering
    \includegraphics[width=0.8\textwidth]{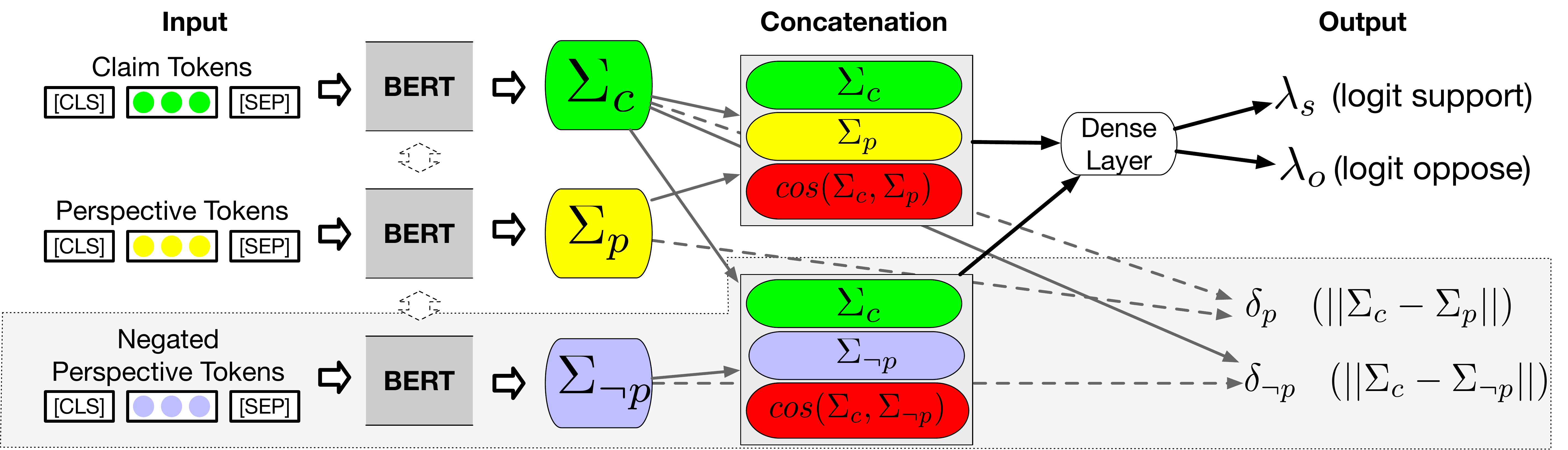}
    \caption{\sys{} network: The components in the light gray area are the ones
        that process the negated perspectives}
    \label{fig:tribid}
\end{figure*}

Current solutions deploy \emph{one} BERT model passing, as input, the claim
followed by the perspective to obtain the latent representation (see, e.g.,
Figure~\ref{fig:bert}). This approach is not ideal for us because we would like
to pass more information as input (the negated perspective), and this might lead
to a string that is too long.  We address this problem by using \emph{multiple}
BERT models that share the same parameters, thus creating a network that is
often labeled as a \emph{siamese} network~\cite{bromley1993signature}.

A schematic view of \sys{} is shown in Figure~\ref{fig:tribid}.  As input,
\sys{} receives the triplet $\langle \claim,\pers,\negpers\rangle$ where
\negpers{} is the negated perspective. We also provide a simpler architecture
where the input is the pair $\langle \claim,\pers\rangle$, which we call
\syspos{}.

In \sys{}, each component of the input triplet is fed to a BERT model that
shares the parameters with the other two models.  The three BERT models compute
latent representations for \claim{}, \pers{}, and \negpers{}, henceforth written
as \claimrep{}, \persrep{}, and \negpersrep{}. In the second stage, the network
concatenates them into a single representation. We experimented with the
concatenation techniques proposed in Sentence-BERT \cite{reimers2019sentence},
InferSent \cite{conneau2018senteval} and Universal Sentence Encoder
\cite{cer2018universal}, namely $(|\claimrep - X|, \claimrep \ast X)$,
$(\claimrep, X, \claimrep \ast X)$, $(\claimrep, X, |\claimrep - X|)$ and
$(\claimrep, X, |\claimrep - X|, \claimrep \ast X)$ where
$X\in\{\persrep,\negpersrep\}$, $|\ldots|$ is the element-wise distance, and
$\ast$ is the element-wise multiplication. We selected $(\claimrep, X,
|\claimrep - X|, \claimrep \ast X)$ as it slightly outperformed the others in
our experiments and further concatenated $cos(\claimrep, X)$ to it because their
similarity is valuable for predicting the stance~\cite{popat2019stancy}.

Finally, the result of the concatenation is passed to a final dense layer that
returns two logits \lpos{} and \lneg{}, which estimate the likelihood of
supporting and opposing stances, respectively.

To train the model, we introduce the following loss function $\loss = \losscos +
\lossent +
\lossdistance$, described below (with \syspos{}, $\loss=\losscos + \lossent$).
The first component \losscos{} is a standard cross entropy loss:
\begin{equation}
    \losscos = -log\left(\frac{exp(\hat{y})}{exp(\lpos) + exp(\lneg)}\right)
\end{equation}
\noindent where $\hat{y}\in \{\lpos,\lneg\}$ is the logit of the true stance.

The second part is the cosine embedding loss:
\begin{equation} \lossent = \left \{ \begin{array}{ll} 1-cos(\claimrep,\persrep)
& \textrm{if } y=1  \\ max(0,cos(\claimrep,\persrep)) & \textrm{if } y=-1  \\
\end{array}\right.  \end{equation}
where $y=1$ if the perspective supports the claim and $-1$ if it is opposed to
it.

The third component $\lossdistance{}$ is added because \losscos{} and \lossent{}
do not take into account the fact that the perspective that supports the claim
should be ``closer'' to the claim than the perspective that opposes it. To this
end, we add a triplet loss~\cite{schroff2015facenet}. Let $\Sigma^+= \persrep$
and $\Sigma^-=\negpersrep$ if the input perspective supports the claim or
$\Sigma^+= \negpersrep$ and $\Sigma^-=\persrep$ otherwise.  Then,
\begin{equation} \lossdistance = max(\gamma + \distsupport - \distoppose, 0)
\end{equation}
\noindent with $\gamma$ being the margin, $\distsupport =|\claimrep -
\Sigma^+|$, and $\distoppose =|\claimrep - \Sigma^-|$ .

\subsection{Stance Classification}
\label{sec:stance}

We can make use of four signals to predict the stance. The first two are the
logits \lpos{} and \lneg{}. The third one is $\distpers = ||\claimrep -
\persrep||$, i.e., the distance between the claim and the input perspective. The
fourth one is $\distnegpers =||\claimrep - \negpersrep||$, i.e., the distance to
the negated perspective. The values of these signals can be combined in
different ways in order to compute a final binary decision. We define two
possible alternative procedures.

The first procedure consists of picking the logit with the highest value as the
final label, e.g., if $\lpos{} > \lneg{}$, then the stance should be support.
In contrast, the second procedure looks at the distance values. In this case, it
chooses the final label depending on which perspective has the closest distance,
i.e., the system should return ``support'' if $\distpers{} < \distnegpers{}$ or
``oppose'' otherwise.

In both cases, the confidence of the model can be quantified by the difference
between the two signals. If difference is at least $\tau$, where $\tau$ is a
given threshold, then we can accept the outcome trusting that the system is
sufficiently confident. Otherwise, we abstain from making a prediction.
Following this principle, we introduce the decision procedures \classlog{} and
\classdist{}, defined as follows:
\begin{equation}
      \classlog=\left\{ \begin{array}{ll}
                                    \support & |\lpos-\lneg|\geq \tau \textrm{
                                        and } \lpos \geq \lneg \\
                                    \oppose &  |\lpos-\lneg|\geq \tau \textrm{
                                        and } \lpos < \lneg \\
                                        \abstain & \textrm{otherwise} \\
                  \end{array}\right.
\end{equation}
\noindent and %
\begin{equation}
      \classdist=\left\{ \begin{array}{ll}
                                    \support & |\distpers -\distnegpers|\geq \tau \textrm{
                                        and } \distpers < \distnegpers \\
                                    \oppose &  |\distpers -\distnegpers|\geq \tau \textrm{
                                        and } \distpers \geq  \distnegpers \\
                                        \abstain & \textrm{otherwise} \\
                  \end{array}\right.
\end{equation}
\noindent where \support{} stands for support, \oppose{} for oppose, and
\abstain{} for abstain.

The advantage of these procedures is that we can select the minimum amount of
acceptable confidence by choosing an appropriate $\tau$. In practice, we can use
a small validation dataset or pick $\tau$ so that at most $X$\% of the data is
excluded.

In case the user is not willing to discard any prediction, then we propose a
third decision procedure where the system never abstains. In essence, our
proposal consists of creating multiple $\classlog$ and $\classdist{}$
classifiers with different $\tau$ which are fed to an ensemble method that makes
the final prediction.

A simple example of an ensemble method is majority voting, but this technique
does not consider latent correlations between the classifiers. To take those
into account, we can rely on weak supervision. In particular, we can use the
state-of-the-art method proposed
by~\citeauthor{fu2020fast}~\shortcite{fu2020fast}, which is called
\emph{FlyingSquid}. As far as we know, methods like the one
of~\citeauthor{fu2020fast} have not yet been considered for stance
classification. We show here that they lead to an improvement of accuracy.

The main goal of \emph{FlyingSquid} is to learn a model that is able to compute
a probabilistic label (which is the stance in our case) with a set of noisy
labeling functions (in our case the \classlog{} and \classdist{} classifiers)
given as input. This method is particularly interesting to us for two reasons:
1) It does not need to access ground truth annotations (which are scarce in our
context), and 2) it can find the optimal model's parameters quickly, without
iterative procedures like gradient descent.

We proceed as follows. First, we create $n$ classifiers \classlogvar{i} and
\classdistvar{j} where $i\in \{\tau^{K}_1,\ldots, \tau^{K}_n\}$ and
$j\in\{\tau^{\Lambda}_1,\ldots,\tau^{\Lambda}_n\}$. For a given pair $\langle
C,P,NP\rangle$, these classifiers produce $2n$ labels that can be either
$\{\support,\oppose,\abstain\}$. These labels form the input for
\emph{FlyingSquid}, which learns a model from the labels' correlations and
return a final label $l\in\{\support,\oppose\}$ for every $\langle
C,P,NP\rangle$.

To recap, we proposed three approaches for stance classification with our neural
model. The first is the \classlog{} classifier, the second is \classdist{},
while the third is a weak supervision model (\emph{FlyingSquid}) built from
multiple \classlog{} and \classdist{} classifiers. The first two classifiers
might abstain if the model is not confident while the third one always returns a
binary output. Henceforth, we refer to them as \syslog{}, \sysdist{}, and
\sysweak{}, respectively.

\subsection{Templates for Automatic Negation}
\label{sec:templates}

To create negated perspectives, we use templates that are encoded as if-then
rules of the form $A \Rightarrow B$. The rules contain instructions on how to
change the text. In case more rules apply to the same perspective, then only the
first application is kept.

For our purpose, the templates should be relatively simple so that they can
applied to large volumes of text and do not capture biases in some datasets.
Moreover, the templates should make only few changes to the text because we
would like to teach the model to pay attention to specific tokens, or
combinations of words, that can potentially change the stance.

With this desiderata in mind, we randomly picked some perspectives from multiple
datasets and negated them by encoding meaningful changes into rules. This
process returned a list of about 60 templates. From this list, we extracted 14
templates which are enough to cover about 90\% of the cases (see
Section~\ref{sec:eval_confidence} for more details about the coverage and the
appendix for the list of all templates). A few examples of templates are shown
below:

\begin{center}
    \footnotesize
    \begin{tabular}{lrcl}
        $P_1:$ &    [X] is/was/... [Y] & $\Rightarrow$ & [X] is/was/... not [Y] \\
        $P_2:$ &    [X] more [Y]       & $\Rightarrow$ & [X] less [Y] \\
        $P_3:$ &    [X] help [Y]       & $\Rightarrow$ & [X] spoil [Y] \\
    \end{tabular}
\end{center}

As we can see from the list, these patterns are fairly simple and mostly reduce to
a strategical position of ``not'' or to replace words with antonyms.

\begin{table}[tb]
    \centering
    \subfloat[Perspectrum (PER)]{
    \footnotesize
    \begin{tabular}{p{2.9em}rllr@{}}
        \toprule
        \textbf{Split}       & Train & Dev  & Test & \textbf{Total} \\ \midrule
        S pairs & 3603  & 1051 & 1471 & 6125           \\
        O pairs        & 3404  & 1045 & 1032 & 5751           \\ \midrule
        \textbf{Total} & 7007  & 2096 & 2773 & 11876          \\ \bottomrule
    \end{tabular}
    \label{tab:statistics_Perspectrum}}
    \\
    \subfloat[IBMCS]{
        \footnotesize
    \begin{tabular}{rrr@{}}
        \toprule
        Train  & Test & \textbf{Total} \\ \midrule
        625   & 700 & 1325           \\
        414   & 655 & 1069           \\ \midrule
        1039  & 1355 & 2394         \\ \bottomrule
    \end{tabular}
    \label{tab:statistics_IBMCS}}
    \caption{Dataset Statistics; S=Support, O=Oppose}
\end{table}

\section{Evaluation}

We tested our approach on the datasets PER~\cite{chen2019seeing} and
IBMCS~\cite{bar-haim-etal-2017-stance}, which are the main datasets previously
used by our competitors. PER is a set of claims and perspectives constructed
from online debate websites while IBMCS is a similar dataset released by IBM.
Statistics on both datasets are in Table~\ref{tab:statistics_Perspectrum}
and~\ref{tab:statistics_IBMCS}. We did not use the datasets proposed in
SemEval-2016, Task 6~\cite{mohammad2016semeval} and SemEval2017, Task
8~\cite{derczynski2017semeval} for the predictions because our work focuses on
a binary classification while these datasets also include additional classes such
as ``neutral'', ``query'', or ``comment''. Extending our method to predict more
than two classes should be seen as future work.

We implemented our approach using PyTorch 1.4.0, and a BERT BASE model with 12
layers, 768 hidden size, and 12 attention heads. We fine-tuned BERT with grid
search optimizing
the $F_1$ on a validation dataset with a learning rates \{1, 2, 3, 4, 5\}
$\times$ {$10^{-5}$}, batch size \{24, 28, 32\}, and the Adam optimizer. For our
experiments, we used a machine with a TitanX GPU with 12GB RAM.

\begin{table}[t]
    \footnotesize
    \centering
    \begin{tabular}{lrrr}
        \toprule
        \begin{tabular}[c]{@{}l@{}}\textbf{Data}\\\textbf{Collection}\end{tabular} & \begin{tabular}[c]{@{}c@{}}\textbf{Cover}\\\textbf{ Cases} \end{tabular} & \begin{tabular}[c]{@{}c@{}}\textbf{Total}\\\textbf{ Cases} \end{tabular} & \begin{tabular}[c]{@{}c@{}}\textbf{Cover}\\\textbf{ Rate} \end{tabular} \\
        \hline
        PER train claim & 6514 & 7007 & 0.930 \\
        PER train perspective & 6270 & 7007 & 0.895 \\
        \hline
        PER test claim & 2618 & 2773 & 0.944 \\
        PER test perspective & 2527 & 2773 & 0.911 \\
        \hline
        PER dev claim & 1866 & 2096 & 0.890 \\
        PER dev perspective & 1881 & 2096 & 0.897 \\
        \hline
        IBMCS train claim & 1039 & 1039 & 1.000 \\
        IBMCS train perspective & 927 & 1039 & 0.892 \\
        \hline
        IBMCS test claim & 1274 & 1355 & 0.940 \\
        IBMCS test perspective & 1192 & 1355 & 0.880 \\
        \hline
        ARC train set & 13878 & 14233 & 0.975 \\
        ARC test set & 3478 & 3559 & 0.977 \\
        \hline
        SemEval2016 train set & 2280 & 2814 & 0.810 \\
        SemEval2016 test set & 1013 & 1249 & 0.811 \\
        \bottomrule
    \end{tabular}
    \caption{Number of perspectives (cases) that can be negated at least by one of our
    templates vs. the total number of perspectives}
    \label{tab:statstemplates}
\end{table}

\begin{table}[tb]
\centering
\footnotesize
\begin{tabular}{lcc}
\toprule
\textbf{Dataset} & \textbf{F1 (Precision, Recall)} & \textbf{Coverage} \\
\hline
PER & 83.54 (82.53, 84.56) & 86.62\% \\
IBMCS & 73.06 (73.80, 72.33) & 86.83\% \\
\bottomrule
\end{tabular}
\caption{Percentage of cases when the logits and distance values agree, and
$F_1$, precision and recall on this subset of cases}
\label{tab:confident_cases}
\end{table}

\begin{table*}[tb]
    \centering
    \footnotesize
    \begin{tabular}{lrrrrrrrrr}
        \toprule
        \textbf{Filtered Percentage} & 10\% & 20\% & 30\% & 40\% & 50\% & 60\% & 70\% & 80\% & 90\% \\
        \hline
        \bertbase{} & 71.94 & 73.98 & 76.44 & 78.40 & 80.61 & 81.51 & 77.89 & 67.10 & 7.14 \\
        \stancy{} & 79.41 & 81.46 & 82.91 & 82.75 & 79.92 & 69.89 & 1.48 & 0.00 & 0.00 \\
        \syslog{} & 82.87 & 85.49 & 86.88 & 87.60 & 87.51 & 85.54 & 85.31 & 88.56 & 86.71 \\
        \sysdist{} & 80.65 & 83.01 & 84.69 & 86.42 & 87.58 & 89.54 & 91.26 & 93.18 & 96.02 \\
        \bottomrule
    \end{tabular}
    \caption{$F_1$ with \syslog{} and \sysdist{} varying the threshold values on
    PER}
    \label{tab:confidences}
\end{table*}

In the following, we first discuss the results using the confidence based
\classlog{} and \classdist{} classifiers (\syslog{} and \sysdist{}). Then, we
study the performance with our weakly supervised approach. Finally, we provide
an analysis on the coverage of the templates for negating the perspectives.

\subsection{Confidence-based Stance Classification}
\label{sec:eval_confidence}

To establish how general our templates are, we have applied them to the text in
PER and IBMCS and in the dataset of SemEval2016, Task 6. Moreover, we also
considered datasets which are used for other NLP tasks, namely ARC
\cite{habernal2018argument, hanselowski2018retrospective}. As we can see in
Table~\ref{tab:statstemplates}, the templates generalize well as they can negate
more than 90\% of the perspectives both in PER and IBMCS and between 81.0\% and
100\% in the other cases. For now, we restrict our analysis to the subset of
perspectives for which there is a negation. In the next section, we will also
consider the (few) remaining cases.

First, it is interesting to look at the percentage of cases when the four
signals, i.e., \lpos{}, \lneg{}, \distpers{}, and \distnegpers{}, agree. If this
occurs for a certain input, then we can interpret it as an hint that the model
is confident about the prediction. For instance, if $\lpos{} > \lneg{}$ and
$\distpers{} < \distnegpers{}$, then we are confident that the output should be
support. Table~\ref{tab:confident_cases} reports the percentage of the input
where the signals agree (column ``Coverage'') and the $F_1$ that we would obtain
if we follow this strategy for deciding the stance. We see that the number of
cases when there is an agreement is large (86\% on PER) and the performance is
fairly high ($F_1$ of about 83.5) and superior to the one that we would obtain
with, e.g. STANCY over the entire dataset (77.76, Table~\ref{tab:eval_results}).
From this, we conclude that this is a rather simple strategy to improve the
performance without discarding many cases. However, notice that with this
approach we cannot choose which is the minimum level of acceptable confidence.
For this, we can use our proposed \syslog{} or \sysdist{}.

In general, if we want to accept only the cases with high confidence, then we
can use either \syslog{} or \sysdist{} with a high $\tau$. In this way, however,
it is likely that we will discard many cases. To study what the tradeoff would
be in our benchmarks, we present the results of an experiment where we pick
$\tau$ such that only $X\%$ of the data will be missed. Two natural baselines
for comparing our performance consist of applying the \classlog{} decision
procedure using the logits returned by \bertbase{} and \stancy{}. In this way,
we can also study the behaviour using other methods. Notice that in this
experiment $\tau$ is chosen independently for each method. This means that the
value of $\tau$ with \stancy{} can be different than the value of $\tau$ with
\bertbase{} when, for instance, $10\%$ of the predictions are filtered out. In
this way, the comparison is fair since it is done on subsets of predictions
which have similar sizes.

\begin{figure}[t]
    \centering
    \includegraphics[width=0.45\textwidth]{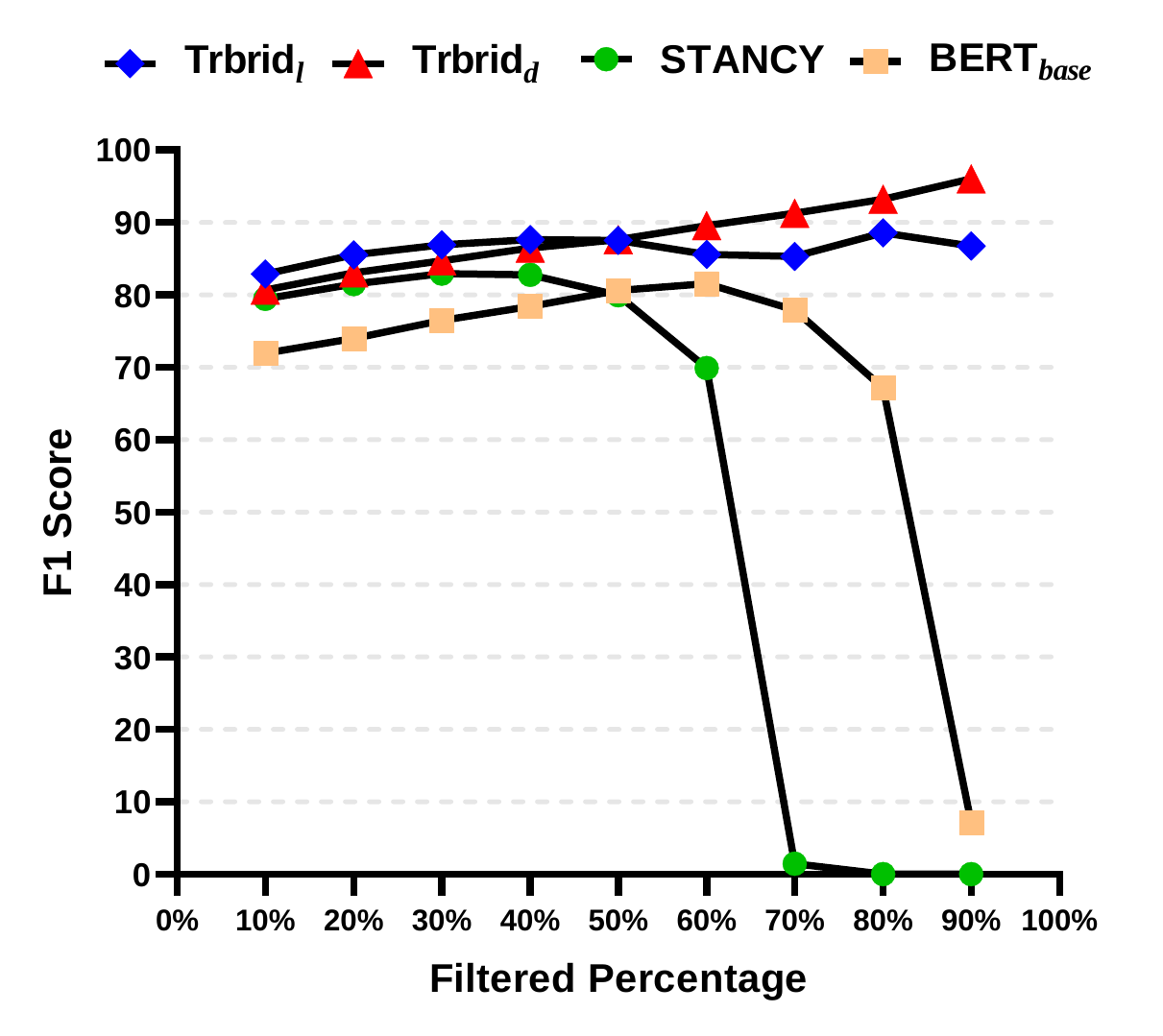}
    \caption{$F_1$ varying $\tau$ with various approaches on PER}
    \label{fig:threshold_graph}
\end{figure}

Table~\ref{tab:confidences} reports the results of this experiments while
Figure~\ref{fig:threshold_graph} plots the same numbers in a graph. As expected,
we observe the that $F_1$ increases as we increase $\tau$.  However, notice that
with \bertbase{} and \stancy{} the $F_1$ drastically decreases after we filter
approximately more than 70-80\% of the cases. That point marks the maximum $F_1$
that we can achieve with those methods. Instead, with our method the $F_1$ keeps
increasing to higher values, which indicates that our system is capable of
producing much higher-quality predictions.

We make \emph{three} main observations. First, with a similar discard rate, our
approach outperforms both \bertbase{} and \stancy{} (the difference is
statistically significant with a p-value of 0.0379 as per paired t-test between
\syslog{} and \stancy{} and 0.0441 between \sysdist{} and \stancy{}). This shows
that our proposed neural architecture, which is trained and used including
negated perspectives, is able to return higher-quality predictions than existing
methods.  Second, \sysdist{} (\classdist{}) can achieve a very high $F_1$; above
95 in the most selective case. However, if we are not willing to sacrifice a
large part of the input, then \syslog{} returns better performance. For
instance, with a discard rate of $30\%$, then \syslog{} returns an $F_1$ of
86.88 vs. 84.69 obtained with \sysdist{}. Finally, it is remarkable that both
\sysdist{} and \syslog{} can achieve a very high accuracy while retaining a
sizable part of the input. We argue that in scenarios such as social media even
if we remove 30-40\% of the available perspectives then we are still left with
enough data for the downstream task (e.g., fact-checking). Clearly, this does
not hold in contexts where all data is needed. In this case, \sysweak{} is more
appropriate.

\begin{table}[b]
    \centering
    \small
    \begin{tabular}{lcc}
        \toprule
        \textbf{Models}  &
        \begin{tabular}[c]{@{}c@{}}\textbf{PER}\\\textbf{ F1 (Prec.,
            Rec.)} \end{tabular} &
            \begin{tabular}[c]{@{}c@{}}\textbf{IBMCS}\\\textbf{ F1 (Prec., Rec.)} \end{tabular} \\
        \hline
        Random& 50.11 & 48.64 \\
        Majority & 34.66 & 34.06 \\
        \hline
        \bertbase{}  & 70.80 (70.50, 71.10) & 63.99 (64.13, 63.86) \\
        \stancy{} & 77.76 (81.14, 74.65) & - \\
        \bertbaseneg{} & 73.83 (65.85, 84.16) & - \\
        \stancyneg{} & 78.13 (70.19, 88.10) & - \\
        \hline
        \syspos{}  & 80.40 (80.33, 80.49) & 70.75 (69.02, 72.57) \\
        \sysweak{} & \textbf{81.35} (81.00, 81.71)  & \textbf{71.16} (71.92, 70.42)  \\
        \hline\hline
        Human & 90.90 (91.30, 90.60) & - \\
        \bottomrule
    \end{tabular}
    \caption{\syspos{} and \sysweak{} vs random baseline (Random), majority
    baseline (Majority), \bertbase{}, \stancy{}, \bertbase{} with negation
(\bertbaseneg{}), \stancy{} with negation (\stancyneg{}), and human performance (Human)}
    \label{tab:eval_results}
\end{table}

\begin{table}[t]
\small
\centering
\begin{tabular}{lc}
\toprule
\textbf{Popular Templates} & \textbf{Coverage} \\
\hline
\textit{[X] is/was/are/were [Y]} & 28.10\% \\
\begin{tabular}[c]{@{}l@{}}\textit{[X] will/would/can/could/shall/}\\\textit{should/may/might/must [Y]}\end{tabular} & 26.28\% \\
\textit{[X] to do (or any verb) [Y]} & 8.18\% \\
\textit{[X] have/has [Y]} & 7.00\% \\
\textit{[X] benefit/help [Y]} & 1.53\% \\
\textit{...} & ...\\
\hline\hline
\textbf{Total} & 91.36\% \\
\bottomrule
\end{tabular}
\caption{Most popular templates on PER. The coverage shows the \% of
    cases where the template matches (for conciseness, we report
only the template's premise)}
\label{tab:templatecoverage}
\end{table}

\subsection{Stance Classification with \sysweak{}}

To exploit the weakly supervised model used in \sysweak{} (\emph{FlyingSquid}),
we created five classifiers with different threshold values. Then, we performed
grid search and feature ablation using the validation dataset and $F_1$ as
metric to optimize. For $\classdist$, we considered threshold values in the
range $[0.01,2]$ while for $\classlog$ the range was $[1,20]$. We then picked
the best performed settings, namely five $\tau$ with the values $\langle 0.01,
0.2, 1.3, 1.5, 1.9\rangle$ for $\classdist$, and five $\classlog$ classifiers
setting $\tau$ with the values $\langle 5, 5.5, 8.5, 11, 13\rangle$. We applied
each classifier to the input and construct a feature vector with $2n$ labels for
every $\langle C,P,NP\rangle$. To train the probabilistic model of
\emph{FlyingSquid}, we used the labels obtained from the claims and perspectives
in the training set.

Table~\ref{tab:eval_results} reports the results obtained with \sysweak{} and
with the simpler variant \syspos{}, which makes no use of negation. Moreover, it
also reports the results with several alternatives. First, we considered
\bertbase{}, \stancy{}, and majority voting as alternative to weak supervision.
Then, we trained additional \bertbase{} and \stancy{} models considering only
the negated perspectives. We used the logits produced by these models and the
ones produced by the models trained with the original perspectives to create 10
$\classlog$ classifiers. In this way, we could evaluate our weak supervision
approach using \bertbase{} and \stancy{} instead of our neural model. We call
these last two baselines \bertbaseneg{} and \stancyneg{}, respectively.

Notice that Table~\ref{tab:eval_results} does not include the results obtained
with the method
by~\citeauthor{schiller2020stance}~\shortcite{schiller2020stance} because it
uses BERT large and external datasets with transfer learning.  Thus, it cannot
be directly compared to our approach. For fairness, we mention that their best
result is a $F_1$ of about 84 on PER. This makes transfer learning a promising
extension of our work, but this deserves a dedicated study. Finally, notice that
if it is not possible to negate the input perspective, then \sysweak{} applies
the fallback strategy of executing \syspos{}. Therefore, the results presented
in Table~\ref{tab:eval_results} were obtained considering the entire testsets.

After looking at Table~\ref{tab:eval_results}, we make \emph{two} main
observations. First, \syspos{} slightly outperforms \stancy{} on PER (80.40 vs
77.76). This means that our strategy of processing the claim and perspective
separately, instead concatenating them is beneficial. We argue that this is
because with our approach BERT receives shorter strings, thus it is able to
produce latent representations of higher quality. Besides, the way we used for
concatenating the representations could also lead to an additional increase of
the performance.

Second, \sysweak{} further outperforms \syspos{}, which was the second best,
with a difference that is statistically significant (p-value of 3.610e-16 with
the McNemar test). Moreover, if we compare the performance of \bertbase{} and
\stancy{} with and without the negated perspectives (\bertbase{} vs.
\bertbaseneg{} and \stancy{} vs. \stancyneg{}), then we observe that the $F_1$
increases also in these cases. This suggests that the strategy of including
negative perspectives and to post-process the output in a weakly supervised
fashion is a viable solution to improve the performance over the entire input.

\subsection{Templates for Negating Perspectives}

\begin{table}[!t]
\centering
\resizebox{\linewidth}{!}{%
\begin{tabular}{
    cccccccc}
\toprule
\textbf{Negation} &
\textbf{Coverage} &
\multicolumn{2}{c}{\textbf{\bertbase{}}} &
\multicolumn{2}{c}{\textbf{\stancy{}}} &
\multicolumn{2}{c}{\textbf{\syspos{}}} \\
& & \#Cases & $F_1$ & \#Cases &
$F_1$ & \#Cases & $F_1$ \\
\hline
AppSuff & \textbf{100\%} & 935 & 76.2 & 974 & 81.3 & 1130 & 85.4 \\
DelNot & 15.58\% & 200 & 51.8 & 232 & 60.7 & 269 & 74.4 \\
Ours & 91.36\% & \textbf{1309} & \textbf{78.0} & \textbf{1332} & \textbf{82.3} &
\textbf{1298} & \textbf{87.3} \\
\bottomrule
\end{tabular}
}
\caption{Comparison about negating the text on PER}
\label{tab:comparison_templates}
\end{table}

Our approach heavily depends on the quality of the templates. For us, a good
template is not necessarily a template that alters the meaning in a way that is
considered optimal by a human. Instead, it is a template that alters the text in
a way that improves stance prediction. Moreover, a good template should not be
too specific so that it can be applied to as much text as possible.

Table~\ref{tab:templatecoverage} reports a list of the most popular templates on
PER. As we can see, a very simple template like the top one matches a large
number of cases. One may wonder what the performance would be if we use instead
one of the other known techniques for negating the text.
Table~\ref{tab:comparison_templates} shows what would happen if we use the
methodologies proposed by
\citeauthor{bilu2015automatic}~\shortcite{bilu2015automatic} and by
\citeauthor{camburu2020make}~\shortcite{camburu2020make}, which are the two most
prominent approaches in the current literature. The first method appends the
suffix ``but this is not true'' while the second removes the token ``not''.
Therefore, we call them ``AppSuff'' and ``DelNot'', respectively.  Since the
goal of negating the perspectives is to recognize dubious predictions, we focus
on the number of ``flipped'' cases, that is the number of cases where the
outcome changes if we pass the negated text. For instance, suppose that the
outcome with perspective $A$ is \support{}.  Then, we expect that if we provide
$\neg A$, then the output is \oppose{}. If this happens, then we count this case
as ``flipped''.

Table~\ref{tab:comparison_templates} reports the number of flipped cases and the
$F_1$ that is obtained on the subset with such cases. Notice that here we
consider only models that have not been trained with negated information to
avoid that they learn some biases. As we can see, the number of flipped cases
and $F_1$ are superior with our templates than with the other two methods. This
shows that negating using templates produces sentences that can be recognized
more easily by BERT. Because of this, BERT can return an opposite prediction in
a larger number of cases and with a better accuracy.

We conclude mentioning a couple of ``too hard'' cases for \sys{}, even with a
high $\tau$. The first relates to the claim ``We should drop the sanctions
against Cuba'' and perspective ``Sanctions are not working''. We suspect that
the problem here is that ``not'' produces a double negation that confuses the
model.  The second is the claim ``Animal testing should be banned'' and
perspective ``Animals do not have rights, therefore it is acceptable to
experiment on them''. In this case, computing the semantics of the perspective
requires some entailment that is likely to be too complex for the model.

\section{Conclusion}

In this paper, we introduced a new method to classify the stance of messages
about a given claim. The main idea is to ``inject'' negated perspectives that
are automatically generated into a BERT-based model so that we can filter out
dubious predictions or to improve the overall accuracy.

If we filter out dubious predictions, then we can improve the performance to a
point where the $F_1$ reaches a human-like level without sacrificing a large
part of the input. We believe that discarding a (small) percentage of the input
is not a major issue in data-intensive environments (e.g., social networks)
where many users express their perspectives. However, if the use case is such
that we must always make a prediction, then we have shown how we can leverage
weak supervision to make a judicious prediction based on the confidence of the
model. Also this approach is competitive against the state-of-the-art on
standard benchmark datasets.

Our work opens the door to several follow-up studies. A natural continuation is
to explore whether we can achieve similar results if we negate the claim instead
of the perspective. Moreover, it is interesting to see whether we can construct
paraphrases instead of negated text and modify the architecture accordingly. If
we increase the number of sequences that we pass as inputs, our ``siamese''
approach may no longer work.  If this occurs, then future work is needed to find
some alternatives. Finally, more sophisticated ways to negate the text may lead
to further improvements.

In general, we believe that critically assessing the output of a BERT model
using negated text is a promising technique to evaluate the model's confidence.
Therefore, it can also bring some improvements in other tasks like sentiment
analysis, entity linking, or word sense disambiguation.

\bibliographystyle{acl_natbib}
\bibliography{references}

\appendix

\clearpage
\section{Templates}

The list of 14 templates used to negate the perspectives is reported below.

\begin{center}
    \small
    \begin{tabular}{rcl}
\toprule
\multicolumn{2}{c}{\textbf{Templates}} \\
\hline
\textit{[X] A [Y]} & $\Rightarrow$ & \textit{[X] A not [Y]}\\
\textit{[X] B [Y]} & $\Rightarrow$ &
\textit{[X] B not [Y]} \\
\textit{[X] C [Y]} & $\Rightarrow$& \textit{[X] not C [Y]} \\
\textit{[X] D [Y]} & $\Rightarrow$& \textit{[X] don't/doesn't D [Y]}\\
\textit{[X] benefit/help [Y]} & $\Rightarrow$& \textit{[X] harm [Y]} \\
\textit{[X] allow [Y]}  & $\Rightarrow$& \textit{[X] disallow [Y]} \\
\textit{[X] not/n’t [Y]}& $\Rightarrow$& \textit{[X] [Y]}\\
\textit{[X] more [Y]} & $\Rightarrow$& \textit{[X] less [Y]}\\
\textit{[X] need [Y]} & $\Rightarrow$& \textit{[X] don't need [Y]}\\
\textit{[X] E [Y]}& $\Rightarrow$& \textit{[X] protect [Y]}\\
\textit{[X] cause [Y]}& $\Rightarrow$& \textit{[X] cause no [Y]}\\
\textit{[X] help [Y]}& $\Rightarrow$& \textit{[X] spoil [Y]}\\
\textit{[X] increase [Y]} & $\Rightarrow$& \textit{[X] decrease [Y]}\\
\textit{[X] everyone [Y]} & $\Rightarrow$& \textit{[X] no one [Y]}\\
\hline

\multicolumn{3}{l}{\textit{A=is/was/are/were}}\\
\multicolumn{3}{l}{\textit{B=will/would/can/could/shall/should/may/might/must}}
\\
\multicolumn{3}{l}{\textit{C=to do (or any verb)}}\\
\multicolumn{3}{l}{\textit{D=have/has}}\\
\multicolumn{3}{l}{\textit{E=hurt/harm/damage}}\\
\bottomrule
\end{tabular}
\end{center}

\end{document}